\documentclass[preprint]{article}
\PassOptionsToPackage{authoryear,square}{natbib}
\usepackage{neurips_2025}
\usepackage[utf8]{inputenc}
\usepackage[T1]{fontenc}
\usepackage[
    colorlinks,
    linkcolor={red!50!black},
    citecolor={blue!50!black},
    urlcolor={blue!80!black}]{hyperref}       % hyperlinks
\usepackage{url}
\usepackage{booktabs}
\usepackage{amsmath,amssymb}
\usepackage{graphicx}
\usepackage{xcolor}
\usepackage{microtype}
\usepackage{pifont}
\newcommand{\xmark}{\textcolor{red}{\ding{55}}}
\newcommand{\cmark}{\textcolor{green!60!black}{\ding{51}}}
\newcommand{\wmark}{\textcolor{orange}{\ding{51}}}

\title{Putnam 2025 Problems in Rocq\\using Opus 4.6 and Rocq-MCP}

\author{
  Guillaume Baudart \\
  IRIF, Université Paris Cité, Inria, CNRS \\
  \And
  Marc Lelarge \\
  DI ENS, PSL University, Inria \\
  \AND
  Tristan Stérin \\
  PRGM DEV
  \And
  Jules Viennot \\
  IRIF, Université Paris Cité, Inria, CNRS \\
}

\begin{document}

\maketitle

\begin{abstract}
We report on an experiment in which Claude Opus~4.6, equipped with a suite of Model Context Protocol (MCP) tools for the Rocq proof assistant, autonomously proved 10 of 12 problems from the 2025 Putnam Mathematical Competition.
The MCP tools, designed with Claude by analyzing logs from a prior experiment on miniF2F-Rocq, encode a \emph{compile-first, interactive-fallback} strategy.
Running on an isolated VM with no internet access, the agent deployed 141 subagents over 17.7 hours of active compute (51.6h wall-clock), consuming approximately 1.9 billion tokens.
All proofs are publicly available.\footnote{\url{https://github.com/LLM4Rocq/Putnam2025-Rocq}}
\end{abstract}

%% ═══════════════════════════════════════════════════════════════════
\section{Introduction}

The Putnam Mathematical Competition is one of the most prestigious undergraduate mathematics contests in North America.
Until recently, LLM-based theorem proving relied primarily on models specialized for a single proof assistant. DeepMind's AlphaProof~\citep{alphaproof} pioneered this formal approach, reaching silver-medal level at the 2024~IMO using reinforcement learning in Lean. Since then, specialized provers, e.g., DeepSeek-Prover~\citep{deepseekprover,deepseekproverv2}, Kimina-Prover~\citep{kimina}, Goedel-Prover~\citep{goedel}, SeedProver~\citep{seedprover}, and Aristotle~\citep{aristotle}, have pushed performance further, with Aristotle achieving gold-medal level at the 2025~IMO. All these systems target Lean and rely on fine-tuning or reinforcement learning.
A striking recent shift is the emergence of \emph{agentic} approaches built on general-purpose frontier models: Numina-Lean-Agent~\citep{numina2025putnam} solved all 12 Putnam 2025 problems in Lean by pairing Claude with an MCP-based tool server.
This pivot from specialized models to tool-augmented frontier agents has an important consequence for proof-assistant diversity.
Unlike fine-tuned provers that are heavily trained on Lean code, general-purpose frontier models have no strong bias toward any single proof language, and their mathematical reasoning is largely decoupled from the target formalism.
The gap between proof assistants therefore narrows considerably.
Our experiment tests this hypothesis on Rocq (formerly Coq), a proof assistant that has received far less attention from the LLM community than Lean.

We present a case study in which Claude Opus~4.6~\citep{claude2025}, orchestrated by Claude Code~\citep{claudecode}, formally verified Putnam 2025 problems in Rocq using \emph{rocq-mcp}~\citep{rocqmcp}, a set of eight MCP tools~\citep{mcp} developed with Claude using feedback from a prior experiment on the miniF2F-Rocq dataset~\citep{minif2f-rocq}.
The Putnam statements were autoformalized from Numina's~\citep{numina2025putnam} and Axiom's~\citep{axiom2025putnam} Lean versions, combined with natural-language statements from the Putnam archive~\citep{putnamarchive}.

\paragraph{Contamination risk.}
The experiment ran on an isolated VM with web search disabled.
Although Lean solutions exist on GitHub, they were released in December 2025, after the model's training cutoff of May 2025.

\paragraph{Outline.}
In Section~\ref{sec:setup}, we describe the design of rocq-mcp tools, motivated by empirical analysis of miniF2F logs.
In Section~\ref{sec:results}, we report quantitative results on Putnam 2025: 10/12 solved, with per-problem breakdowns.
In Section~\ref{sec:analysis}, we use Claude to perform an in-depth analysis of the experiment logs, covering tool use, multi-agent orchestration, scaling, and failure modes.

%% ═══════════════════════════════════════════════════════════════════
\section{Background and Experimental Setup}
\label{sec:setup}

\subsection{The rocq-mcp Toolchain}

The rocq-mcp server~\citep{rocqmcp} exposes eight MCP tools organized in two tiers.

\textbf{Compilation tools} (require only \texttt{coqc}):
\begin{itemize}
  \item \textbf{rocq\_compile}: Full-file compilation with structured error reporting, including source-line annotations and caret underlines for error positions.
  This is the primary verification method: 81\% of miniF2F proofs succeeded via direct compilation alone.
  \item \textbf{rocq\_verify}: Since LLM-generated code cannot be trusted, the proof is wrapped inside a Rocq module, isolating it from the rest of the file. The tool then attempts to prove the original theorem statement by applying the theorem proved inside the module. This sandboxing prevents the LLM from silently redefining the problem statement or using \texttt{Admitted}. A whitelist of standard axioms filters out proofs that rely on custom axioms.
  \item \textbf{rocq\_auto\_solve}: Attempts standard automation tactics (\texttt{auto}, \texttt{lia}, \texttt{lra}, \texttt{ring}, \texttt{field}, etc.) as a quick check before manual construction.
\end{itemize}

\textbf{Interactive tools} (require \texttt{pet} from coq-lsp~\citep{coqlsp}):
\begin{itemize}
  \item \textbf{rocq\_query}: Run \texttt{Search}, \texttt{Check}, \texttt{Print}, or \texttt{About} commands for library exploration.
  \item \textbf{rocq\_step} / \textbf{rocq\_step\_multi}: Execute tactics interactively; \texttt{step\_multi} tests up to 20 tactics in parallel without advancing the session.
  \item \textbf{rocq\_toc} / \textbf{rocq\_notations}: File structure and notation resolution queries.
\end{itemize}

These tools encode a \emph{compile-first, interactive-fallback} strategy identified through miniF2F log analysis: the agent writes a complete proof file, compiles it, iterates on errors, and only falls back to interactive stepping for debugging specific subgoals.

\subsection{Tool Design from miniF2F Analysis}

Prior to the Putnam experiment, we ran Claude Opus~4.6 on the miniF2F-Rocq benchmark~\citep{minif2f-rocq} (244 test theorems) with only the Rocq compiler as a tool.
Of the 244~theorems, 198 (81\%) were correctly proved: the agent wrote a complete proof file that compiled and passed manual verification.
The remaining 46~failures fell into categories that each motivated a specific tool:

\begin{itemize}
  \item \emph{Invalid proofs} (20/46): 14~proofs used \texttt{Admitted} and 6~exploited type redefinition to bypass the theorem statement.
  These compiled without errors but were mathematically invalid, motivating a sandboxing mechanism and an axiom whitelist in \texttt{rocq\_verify}.
  \item \emph{Incomplete proofs} (15/46): Proofs with open subgoals that the agent could not detect from compiler output alone, motivating the interactive \texttt{rocq\_step} tool for goal inspection.
  \item \emph{Other failures} (11/46): Timeouts, type errors, and import issues.
  Across all failures, the agent struggled to locate errors from raw \texttt{coqc} output, often compiling entire \texttt{.v} files just to run \texttt{Search} commands, motivating \texttt{rocq\_query}.
  Scope ambiguity (e.g., \texttt{+} in \texttt{nat\_scope} vs \texttt{Z\_scope}) caused silent type errors, motivating \texttt{rocq\_notations}.
\end{itemize}

\subsection{Experimental Protocol}
\label{sec:protocol}

The Putnam 2025 problems were autoformalized from Numina's~\citep{numina2025putnam} and Axiom's~\citep{axiom2025putnam} Lean versions, combined with natural-language statements from the Putnam archive~\citep{putnamarchive}.
Each problem is a standalone \texttt{.v} file containing definitions and a \texttt{Theorem putnam\_2025\_XX : ... Admitted.} stub.

The experiment ran on a Docker container with Rocq~9.0, the rocq-mcp server, and large math libaries (Coquelicot~\citep{coquelicot}, math-comp~\citep{mathcomp}, and math-comp-analysis~\citep{mathcompanalysis}) installed.
Web search was disabled.
The user prompt was:

\begin{quote}
\emph{Launch an agent team with an expert mathematician, an expert computer scientist, an expert in formal verification (Coq/Rocq), and a devil's advocate.
Your goal is to solve the 12 theorems in this directory.}
\end{quote}

Claude Code (orchestrating Claude Opus~4.6) ran autonomously for approximately 3~days, with human intervention limited to resuming sessions after container crashes and rate-limit pauses.

%% ═══════════════════════════════════════════════════════════════════
\section{Results}
\label{sec:results}

\begin{table}[t]
\caption{Per-problem results. Lines = lines of Rocq code. Time = wall-clock from experiment start to successful compilation.
\wmark~= exploited a loophole in the A3 formalization (later fixed by the agent).
Bold \textbf{None} = fully constructive (zero axioms).}
\label{tab:results}
\centering
\small
\begin{tabular}{lcrrcl}
\toprule
\textbf{Problem} & \textbf{Status} & \textbf{Lines} & \textbf{Time} & \textbf{Axioms} & \textbf{Domain} \\
\midrule
A1 & \cmark & 305 & 51m & classical & Number theory \\
A2 & \cmark & 308 & 1h 33m & Reals & Real analysis \\
A3 & \wmark & 110 & 1h 07m & \textbf{None} & Combinatorics \\
A4 & \cmark & 531 & 4h 58m & Reals & Linear algebra \\
A5 & \xmark & 2294 & -- & -- & Enum.\ combinatorics \\
A6 & \cmark & 897 & 19h 56m & \textbf{None} & Number theory \\
B1 & \cmark & 570 & 1h 48m & Reals & Geometry \\
B2 & \cmark & 513 & 5h 13m & Reals & Real analysis \\
B3 & \cmark & 439 & 15h 21m & \textbf{None} & Number theory \\
B4 & \cmark & 414 & 3h 46m & \textbf{None} & Combinatorics \\
B5 & \cmark & 1455 & 46h 01m & \textbf{None} & Number theory \\
B6 & \xmark & 1160 & -- & -- & Analysis \\
\bottomrule
\end{tabular}
\end{table}

The agent solved 10 of 12 problems, producing a total of 5,542 lines of verified Rocq code (Table~\ref{tab:results}).
However, the A3 proof exploited a loophole in the problem formalization rather than proving the intended mathematical statement; it was later fixed by the agent in a separate session (see Section~\ref{sec:discussion}).
Five proofs are fully constructive (A3, A6, B3, B4, B5): they use no axioms.
The remaining solved proofs use standard mathematical axioms (classical logic, Dedekind Reals).

\begin{figure}[t]
\centering
\includegraphics[width=\linewidth]{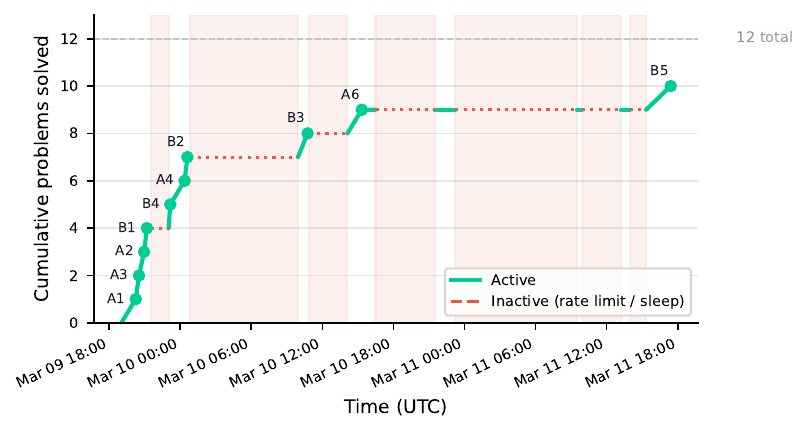}
\caption{Cumulative problems solved over wall-clock time. Solid green segments indicate active compute; dashed red segments indicate inactivity (API rate limits or user absence). Shaded regions mark the 7 detected gaps ($>$30\,min). The first 4 problems were solved within 2 hours; B5 was not solved until 46 hours into the experiment.}
\label{fig:timeline}
\end{figure}

\begin{figure}[t]
\centering
\includegraphics[width=\linewidth]{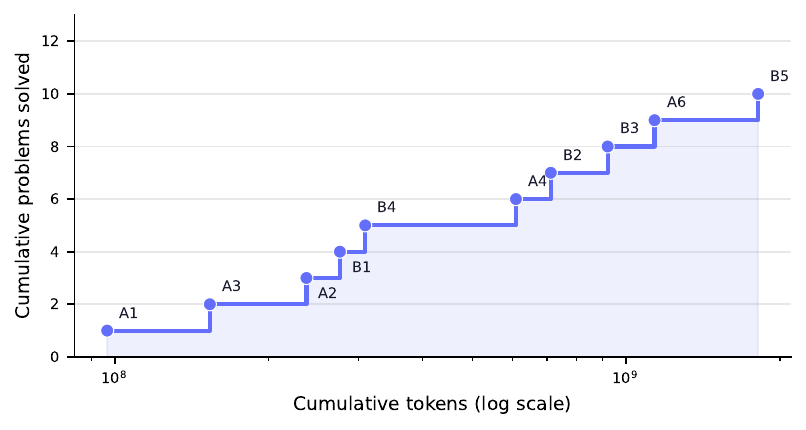}
\caption{Cumulative problems solved vs.\ cumulative tokens consumed (log scale). The first 5 problems required $\sim$100M tokens; the last 5 required $\sim$10$\times$ more, illustrating sharp diminishing returns on harder problems.}
\label{fig:tokens}
\end{figure}

%% ═══════════════════════════════════════════════════════════════════
\section{Analysis}
\label{sec:analysis}

\subsection{How effective are the MCP tools?}
\label{sec:rq1}

Across the experiment, the agent made 12,427 tool calls.
The MCP tools accounted for approximately 30\% of all calls, with \texttt{rocq\_compile} being the dominant verification method (Table~\ref{tab:tools}).

\begin{table}[t]
\caption{MCP tool usage.}
\label{tab:tools}
\centering
\small
\begin{tabular}{lrl}
\toprule
\textbf{Tool} & \textbf{Calls} & \textbf{Role} \\
\midrule
\texttt{rocq\_compile} & $\sim$3,100 & Compilation \\
\texttt{rocq\_verify} & $\sim$120 & Axiom checking \\
\texttt{rocq\_auto\_solve} & $\sim$20 & Quick automation \\
\texttt{rocq\_query} & $\sim$80 & Library search \\
\texttt{rocq\_step} & $\sim$450 & Interactive debug \\
\bottomrule
\end{tabular}
\end{table}

\textbf{Compile-first confirmed.}
Every successful proof was ultimately verified via \texttt{rocq\_compile} or direct \texttt{coqc} invocation.
The compile-iterate loop (write proof $\to$ compile $\to$ fix errors $\to$ repeat) was the universal workflow.
Across all problems, compile success rates ranged from 46\% (A1) to 58\% (A3), with a median of $\sim$50\%, meaning roughly half of compilation attempts produced errors that the agent then fixed.

\textbf{\texttt{rocq\_auto\_solve} is insufficient for Putnam.}
Standard automation never solved a Putnam problem: these require non-trivial mathematical insight, not tactic search.
However, \texttt{rocq\_auto\_solve} remains useful for simpler problems because it provides a fast, zero-cost baseline before engaging the LLM: it solves 53/244 (22\%) of miniF2F-Rocq test problems.

\textbf{All proofs were double-checked with \texttt{rocq\_verify}.}
All 10 solved proofs pass verification with only standard axioms (classical logic, Dedekind Reals, functional extensionality).
However, the module sandbox caused false negatives on two problems (A3, B1) where custom inductive types could not unify across the module boundary.
A two-phase strategy was later added to rocq-mcp~\citep{rocqmcp} to handle this case.

\textbf{Interactive tools see targeted use.}
\texttt{rocq\_step} was used almost exclusively on the three hardest problems (A5, B5, B6), where the agent needed to understand specific proof states.
For the other nine problems, compile-first sufficed.

\subsection{How does the multi-agent architecture perform?}
\label{sec:rq2}

The orchestrator launched 4 parallel teams of 3 problems each, then spawned a total of 141 subagents over the experiment.
The architecture was strictly two-level: one orchestrator managing worker subagents (no sub-subagent spawning observed).

\textbf{Initial burst.}
The parallel dispatch was highly effective: 4 problems solved in 1.8h, 7 in 5.2h (Figure~\ref{fig:timeline}).

\textbf{Specialized follow-up.}
After the initial teams completed, the orchestrator spawned targeted agents classified by role (Table~\ref{tab:roles}).
Lemma Provers (55 agents, 738M tokens) dominated, reflecting the proof-by-subgoal strategy: once an overall structure was established, the hard work was proving specific lemmas.
Bug Fixers (36 agents) handled compile-error repair, while Proof Completers (13 agents) filled in incomplete proof skeletons.
Verifiers (15 agents) were used to check axiom-freeness and eliminate unnecessary \texttt{Admitted} stubs.

\begin{table}[t]
\caption{Subagent roles. Roles are inferred from each agent's initial prompt.}
\label{tab:roles}
\centering
\small
\begin{tabular}{lrrr}
\toprule
\textbf{Role} & \textbf{Agents} & \textbf{Tokens} & \textbf{Tool calls} \\
\midrule
Lemma Prover   & 55 & 738M & 4,845 \\
Bug Fixer      & 36 & 538M & 3,409 \\
General        & 21 & 253M & 1,588 \\
Verifier       & 15 &  65M &   639 \\
Proof Completer & 13 & 185M & 1,257 \\
Compiler       &  1 &   1M &    14 \\
\midrule
\textbf{Total} & \textbf{141} & \textbf{1,780M} & \textbf{11,752} \\
\bottomrule
\end{tabular}
\end{table}

\textbf{Diminishing returns on hard problems.}
B5 received 82+ subagents and B6 received 91, yet many agents attempted the same core mathematical difficulty without fundamentally new approaches.
An early-termination strategy that detects when multiple agents fail at the same bottleneck could save substantial resources.

\begin{figure}[t]
\centering
\includegraphics[width=\linewidth]{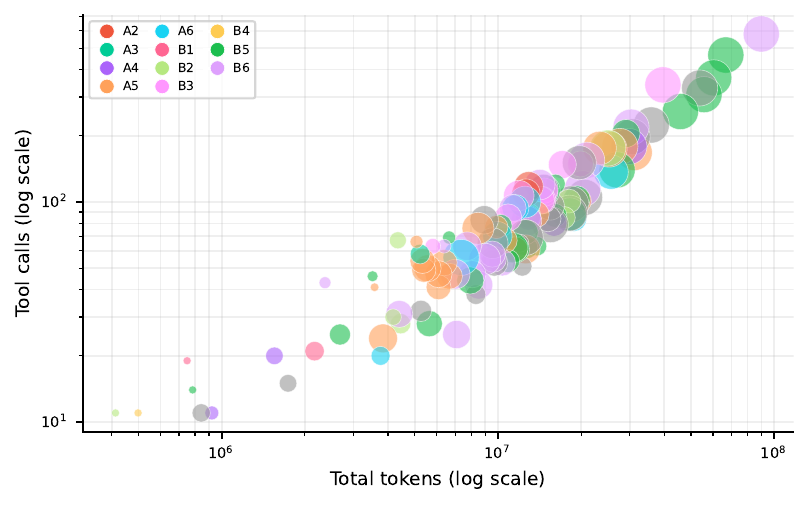}
\caption{Subagent efficiency. Each bubble represents one subagent (141 total, filtered to those with $>$10 tool calls). Position: tokens consumed vs.\ tool calls (both log scale). Size: duration. Color: problem assignment.}
\label{fig:bubble}
\end{figure}

\textbf{Mathematical hints matter.}
Each subagent received multi-paragraph mathematical proof strategies in its prompt.
These hints (e.g., the two-region strategy for A2's upper bound, the equilateral triangle construction for B1) were critical for directing the search.
Notably, these hints were generated by the orchestrator LLM itself, not by a human.

\subsection{How do costs scale with problem difficulty?}
\label{sec:rq3}

The experiment consumed approximately 1.9 billion tokens at an estimated API cost of \$1,760 (Table~\ref{tab:tokens}).

\begin{table}[t]
\caption{Token economics.}
\label{tab:tokens}
\centering
\small
\begin{tabular}{lrrrr}
\toprule
\textbf{Category} & \textbf{Tokens} & \textbf{\%} & \textbf{Cost} & \textbf{\% Cost} \\
\midrule
Input & $<$1M & 0.0\% & \$1 & 0.1\% \\
Output & 16M & 0.9\% & \$406 & 23.1\% \\
Cache creation & 72M & 3.8\% & \$448 & 25.5\% \\
Cache read & 1,809M & 95.4\% & \$905 & 51.4\% \\
\midrule
\textbf{Total} & \textbf{1,897M} & \textbf{100\%} & \textbf{\$1,760} & \textbf{100\%} \\
\bottomrule
\end{tabular}
\end{table}

\textbf{Output tokens dominate cost.}
Despite being only 0.9\% of token volume, output tokens account for 23.1\% of cost at \$25/M, reflecting hundreds of lines of proof code and extended mathematical reasoning per problem.
Cache reads account for the largest share of cost (51.4\%) despite their low per-token price (\$0.50/M), simply due to volume: 95.4\% of all tokens are cache reads.

\textbf{Sharp diminishing returns.}
The first 5 problems consumed a small fraction of total tokens; the remaining 5 consumed the vast majority (Figure~\ref{fig:tokens}).
Table~\ref{tab:scaling} details the budget distribution across difficulty groups.

\begin{table}[t]
\caption{Scaling behavior by problem difficulty. The unsolved problems consumed the most resources.}
\label{tab:scaling}
\centering
\small
\begin{tabular}{lrrr}
\toprule
\textbf{Group} & \textbf{Problems} & \textbf{Active time} & \textbf{Est.\ tokens} \\
\midrule
Easy (A1, A2, A3, B4) & 4 & $<$ 2h & $\sim$100M \\
Medium (A4, B1, B2, B3) & 4 & $\sim$5h & $\sim$400M \\
Hard (A6, B5) & 2 & $\sim$11h & $\sim$600M \\
Unsolved (A5, B6) & 2 & -- & $\sim$800M \\
\bottomrule
\end{tabular}
\end{table}

\textbf{Active time vs.\ wall-clock.}
Only 17.7h of the 51.6h wall-clock was active compute.
The remaining 33.8h consisted of 7 inactivity gaps caused by API rate limits and user absence.

%% ═══════════════════════════════════════════════════════════════════
\section{Discussion}
\label{sec:discussion}

\textbf{The compile-first paradigm.}
On the Putnam 2025 problems considered here, the compile-first strategy was sufficient: the compiler provides precise error messages, and the agent can write and revise complete proof files in a single pass.
Interactive stepping was only useful for debugging specific subgoals, not for proof discovery.
This aligns with our miniF2F findings: 81\% of proofs succeeded without any interactive tools.
Whether this holds for more complex formalization tasks involving large libraries remains an open question.

\textbf{Cost efficiency.}
Prompt caching reduced costs by $5.6\times$.
The sharp scaling wall, where unsolved problems consume disproportionate resources, suggests that better difficulty estimation and early termination could significantly improve cost efficiency.

\textbf{Formalization quality matters: the A3 case.}
Problem~A3 is a combinatorial game: two players alternate moves and the player with no legal move loses.
The initial autoformalization modeled strategies as partial functions (\texttt{option state}).
This encoding had a loophole: the strategy ``never move'' is vacuously compatible with any game, so the agent proved Bob wins with a mathematically trivial argument in 1\,h\,07\,m.
When asked to close this loophole, the agent spent ${\sim}$9\,h attempting to prove the strengthened statement before concluding the encoding was fundamentally flawed.
We then pointed the agent to an alternative formalization from AxiomProver~\citep{axiom2025putnam}, where the current player must exhibit a valid move, which closes the loophole by construction.
The agent translated the formalization from Lean to Rocq and found a correct, fully constructive proof in ${\sim}$1\,h.
This episode illustrates both a strength and a risk of the agentic approach: agents can detect and exploit formalization gaps, but closing them may require human guidance on the problem encoding.

%% ═══════════════════════════════════════════════════════════════════
\section{Conclusion}

Claude Opus~4.6, equipped with empirically designed MCP tools and a compile-first multi-agent workflow, solved 10 of 12 Putnam 2025 problems in Rocq.
This result, achieved with a proof assistant that has received far less LLM attention than Lean, suggests that tool-augmented frontier agents can operate effectively across proof systems without language-specific training.
The A3 episode highlights that formalization quality is a first-class concern in agentic proving: agents can exploit gaps in problem encodings, and closing those gaps sometimes requires human guidance.
All proofs and the rocq-mcp toolchain are publicly available~\citep{putnam2025rocq,rocqmcp}.

\bibliographystyle{plainnat}
\bibliography{main}

\end{document}